\documentclass{article}

\usepackage{arxiv}
\usepackage{cite}
\usepackage{amsmath,amssymb,amsfonts}
\usepackage{algorithmic}
\usepackage{graphicx}
\usepackage{textcomp}
\usepackage{booktabs}
\usepackage{svg}
\usepackage{flushend} 

\usepackage{xcolor}
\usepackage[bookmarks=false]{hyperref}

\usepackage{multicol}

\def\BibTeX{{\rm B\kern-.05em{\sc i\kern-.025em b}\kern-.08em
    T\kern-.1667em\lower.7ex\hbox{E}\kern-.125emX}}

\begin{document}
\newcommand{\da}{DA}
\newcommand{\ccmvpop}{CCMVPOP}

\title{Fast Hyperparameter Tuning for Ising Machines}



\author{ \href{https://orcid.org/0000-0002-5777-7756}{\includegraphics[scale=0.06]{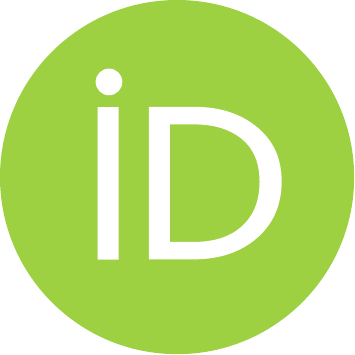}\hspace{1mm}Matthieu~Parizy}\thanks{First author also affiliated with the Department of Computer Science and Communications Engineering of Waseda University. matthieu.parizy@togawa.cs.waseda.ac.jp.} \\
	Fujitsu Research\\
	Fujitsu LTD.\\
	Kawasaki, Japan \\
	\texttt{parizy.matthieu@fujitsu.com} \\
	\And
	Norihiro ~Kakuko \\
	Fujitsu Research\\
	Fujitsu LTD.\\
        Kawasaki, Japan\\
	\texttt{kakuko.norihiro@fujitsu.com} \\
	\AND
	    \href{https://orcid.org/0000-0003-3400-3587}{\includegraphics[scale=0.06]{orcid.pdf}\hspace{1mm}Nozomu ~Togawa} \\
	Department of Computer Science and Communications Engineering \\
	Waseda University \\
        Tokyo, Japan\\
	\texttt{togawa@togawa.cs.waseda.ac.jp} \\
}

\maketitle

\begin{abstract}
In this paper, we propose a novel technique to accelerate Ising machines hyperparameter tuning.
Firstly, we define Ising machine performance and explain the goal of hyperparameter tuning in regard to this performance definition.
Secondly, we compare well-known hyperparameter tuning techniques, namely random sampling and Tree-structured Parzen Estimator (TPE) on different combinatorial optimization problems.
Thirdly, we propose a new convergence acceleration method for TPE which we call ``FastConvergence".
It aims at limiting the number of required TPE trials to reach best performing hyperparameter values combination. 
We compare FastConvergence to previously mentioned well-known hyperparameter tuning techniques to show its effectiveness.
For experiments, well-known Travel Salesman Problem (TSP) and Quadratic Assignment Problem (QAP) instances are used as input. 
The Ising machine used is Fujitsu's third generation Digital Annealer (DA).
Results show, in most cases, FastConvergence can reach similar results to TPE alone within less than half the number of trials.
\end{abstract}


\section{Introduction}
Over the past 10 years, quantum-based Ising machines \cite{dwave_nature} and non-quantum based Ising machines \cite{Matsbuara-DA-General}, \cite{Toshiba_SBM}, \cite{hitachi_cmos_annealing} gave promising results when solving certain kinds of combinatorial problems formalized as quadratic unconstrained binary optimization (QUBO) form. 

Yet their limits have been shown on problems having certain kinds of constraints \cite{my_qkp}. 
To overcome such limitations, hybrid software-hardware technologies such as Fujitsu's third generation Digital Annealer (\da) \cite{da3}, \cite{digital_annealer_homepage} have recently emerged. 
\da{} is now an Ising machine-software system which can handle Binary Quadratic Programs (BQP), expressed using a limited set of constraints, in contrast to conventional Ising machines  \cite{dwave_nature}, \cite{Matsbuara-DA-General}, \cite{Toshiba_SBM}, \cite{hitachi_cmos_annealing} which can only handle QUBO.

Cited Ising machines have a varying number of hyperparameters and tuning them is a time and money consuming task as most machines' usage price are significant.
Popular automated tuning techniques are random sampling, where hyperparameters values are picked randomly, grid-search where predefined combinations of hyperparameters are explored, and manual search done by experts.

A more sophisticated tuning approach which takes into account past results to find the best hyperparameter values is Tree-structured Parzen Estimator (TPE), which is a sequential model-based optimization approach originally created and shown to be highly efficient for neural network hyperparameter tuning \cite{tpe_original}. 
TPE has recently been shown to be efficient for QUBO penalty coefficient tuning, before said QUBO is input in an Ising machine\cite{tpe_ising_example1, tpe_ising_example2, tpe_ising_example3, tpe_ising_example4, tpe_ising_example5} but not for Ising machine hyperparameters themselves.
Tuning of QUBO penalty coefficient is an extensive topic and area of research. 
The reason for its popularity is QUBO penalty coefficient is often the most critical parameter to improve performance.
In fact, this coefficient tuning will result in higher performance across all Ising machines \cite{verma2020penalty,mayowa2022penalty, marcos2022penalty}.
However tuning of hyperparameters of Ising machine themselves is not covered as extensively, thus our motivation to research black-box tuning approach such as random sampling or TPE.

In this paper, after analyzing tuning performance of random sampling and TPE alone, we propose a new convergence acceleration method for TPE we call FastConvergence. 
It aims at limiting the number of required TPE trials to reach best performing hyperparameter values combination. 
We compare FastConvergence to previously mentioned well-known hyperparameter tuning techniques to show its effectiveness.
Our main contributions are: 
\begin{itemize}
    \item We show that TPE converges to better parameters than random sampling and how much improvement it can yield compared to default parameters.
    \item We propose a method called FastConvergence, which allows TPE to converge to better parameters faster than TPE alone.
\end{itemize}

In Section \ref{sec:hyper_param_imp}, we will define Ising machine performance and how to evaluate Ising machines hyperparameters in regard to said performance.
In Section \ref{sec:blackbox_tuning_techniques}, we explain what tuning techniques can be used as baselines and what kind of objective they should target.
In Section \ref{sec:TPE_accel}, we describe our main contribution which aims at accelerating TPE convergence.
In Section \ref{sec:experiments}, we show the efficiency of our proposed method using well-known combinatorial optimization problems: Travel Salesman Problem(TSP) and Quadratic Assignment Problem (QAP). 
Finally Section \ref{sec:conclusion}, gives several concluding remarks.

\section{Evaluating Ising Machine Hyperparameters}\label{sec:hyper_param_imp}
To illustrate the importance of hyperparameter tuning, we must first define Ising machines performance and what problem they solve.
BQP solved by DA are defined as follows:
\begin{alignat}{2}
\label{eq:bqp_obj}
    \text{minimize}\quad & E(x) = x^TQ_{obj}x \\
    \label{eq:bqp_pen_qubo}
    \text{subject to}  \quad & x^TQ_{pen}x =0\\
    \label{eq:bqp_ineq}
    & W_i x \leq C_i, \quad \forall i
\end{alignat}
\noindent where $x=(x_1, \dots,x_m)$ is an $m$-dimensional vector of binary variables. 
$Q_{obj}$, an $m \times m$ matrix called a QUBO matrix, represents the objective function we want to minimize. 
$Q_{pen}$ is an $m \times m$ QUBO constraint to respect, if any. 
$W_i$ the linear inequality  constraint $i$ weights for each binary variable, and $C_i$ the constant for linear inequality constraint $i$, if any.

DA's objective is thus to find the vector $x$ which gives the lowest value for $E(x)$ described in \eqref{eq:bqp_obj} which respects \eqref{eq:bqp_pen_qubo} and \eqref{eq:bqp_ineq} if \eqref{eq:bqp_pen_qubo} and/or \eqref{eq:bqp_ineq} are present.

In our paper, we will focus on BQPs difficult enough so that their optimal solutions cannot be found within the set experiment DA run time $T$. 
Thus hyperparameters performance means the set of hyperparameters which give the lowest $E$ found at $T$ respecting the constraints, which we will call $E_{min}$ hereinafter.
In this paper, we do not discuss trade-off of hyperparameters leading to a better $E(x)$ within a smallest time $t_1 < T$ but have a worse $E(x)$ than others past $t_2$ where $t_1 < t_2 \leq T$.

We define $GAP$ between $E_{min}$ and another hyperparameter set $p$ tried, leading to worse $E_p$ as follows:
\begin{equation}\label{eq:gap}
    GAP = \frac{E_p - E_{min}}{|E_{min}|}
\end{equation}
\noindent where $E_p$ is the lowest $E$ found for given hyperparameter set $p$.
Thus, the best hyperparameter tuning method will lead to the best $p$ with a $GAP_p$ value of $0$, which corresponds to $E_{min}$, and other methods will lead to worse $p$ which will lead to $GAP$ values strictly positive.

\section{Black-box Hyperparameter Tuning Techniques}\label{sec:blackbox_tuning_techniques}
\begin{figure}[tb] 
\centering
\includegraphics[width=0.9\columnwidth]{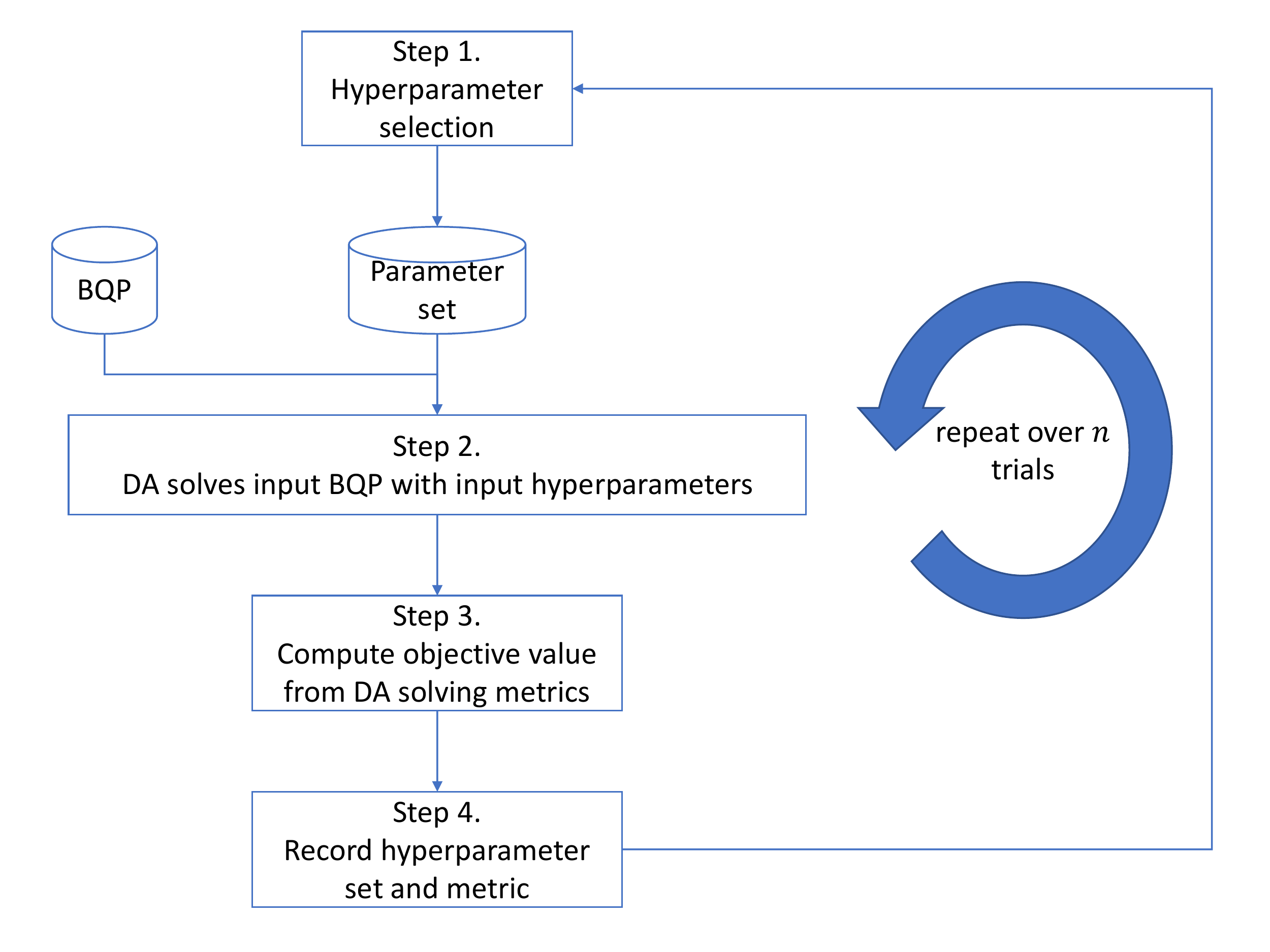}
\caption{DA Black-box tuning method}
\label{fig:tuning_flow}
\end{figure}

Black-box hyperparameter tuning techniques consist in picking a set of hyperparameters, input this set in an objective black-box function which should be minimized or maximized, record said objective value for the tried hyperparameter set and repeat the process until satisfied. 
In our case, the black-box is the DA, which is fed DA hyperparameters values and a given fixed BQP for which the hyperparameters are tuned for.
What we measure are metrics related to finding the lowest $E(x)$ for a given BQP for a given hyperparameter values set.
Metrics are described in Section. \ref{sec:obj_comp}.
This tuning flow is illustrated in Fig. \ref{fig:tuning_flow}.

The hyperparameter selection techniques (step 1 in Fig. \ref{fig:tuning_flow}) we use as baselines for comparison are  random sampling and TPE. 

\subsection{Baseline1: Random Sampling}\label{sec:random_samping}
Random sampling consists in randomly picking combination of hyperparameters within a given range $r$ for each hyperparameter following a set distribution. 
$r_{min}$ is the minimum value allowed for a given parameter and $r_{max}$ its maximum. We thus have $r = r_{max} - r_{min}$ with $r_{max} > r_{min} \geq 0$.
Every time a random combination of hyperparameters is chosen, it is evaluated using DA on the given BQP (step 2) then the objective from DA solving metrics is computed (step 3). 
Finally, the hyperparameter set and its corresponding objective value are recorded during step 4.
After doing $n$ trials, the set of parameters $p$ giving the lowest $E_p$ is chosen as the best parameter set for the hyperparameter selection technique.

\subsection{Baseline2: Tree-structured Parzen Estimator}\label{sec:tpe}
Tree-structured Parzen Estimator (TPE) is a Bayesian method whose main difference with random sampling is that, in order to determine the next hyperparameter values to try (step 1), it will consider past hyperparameter trials recorded during step 4.
Those past trials are used to constitute a surrogate model to the objective to minimize which is differentiable and thus easier to minimize than the original objective function.
This surrogate model represents prior probability distributions $p(x|y)$ and $p(y)$ where $y$ is the expected objective value given hyperparameter value $x$ using  tree-structured adaptive Parzen estimators.
$p(x|y)$ is modeled using one Gaussian mixture model $l(x)$ to the set of hyperparameter values corresponding to the best objective values and another Gaussian mixture model $g(x)$ for the remaining parameter values. TPE selects the hyperparameter value $x$ which maximizes the ratio $l(x)/g(x)$.
TPE is used in hyperparameter optimization frameworks such as Hyperopt \cite{hyperopt} and Optuna \cite{optuna}.
In this paper we use Optuna for our experiments.

\subsection{Objective Computation}\label{sec:obj_comp}
For objective computation (step 3 in Fig. \ref{fig:tuning_flow}), we propose to calculate it as follows:
\begin{equation}\label{eq:obj}
    O(p) = t \times E_p + T_{E_p} 
\end{equation}
\noindent where $O(p)$ is the objective value for given hyperparameter set $p$, $E_p$ is the lowest $E$ found using $p$ during the DA experiment run time $T$, $T_{E_p}$ is the time at which solution corresponding to $E_p$ was found, and $t$ is a coefficient we set to $T$. 
The reason we do not simply record $E_p$ is to differentiate two sets of parameters leading to the same $E_p$.
By adding $T_{E_p}$, the hyperparameter set which leads to finding a solution corresponding to $E_p$ the fastest is prioritized. 
Differentiating parameter set leading to same $E_p$ is only useful for TPE based methods as it models the relation between hyperparameters and objective value to gradually converge to the best possible $p$. It is also only useful for BQP where $E_p$ ties happen often.

\section{Tree-structured Parzen Estimator acceleration for Ising Machines}\label{sec:TPE_accel}
\begin{figure}[tb] 
\centering
\includegraphics[width=0.9\columnwidth]{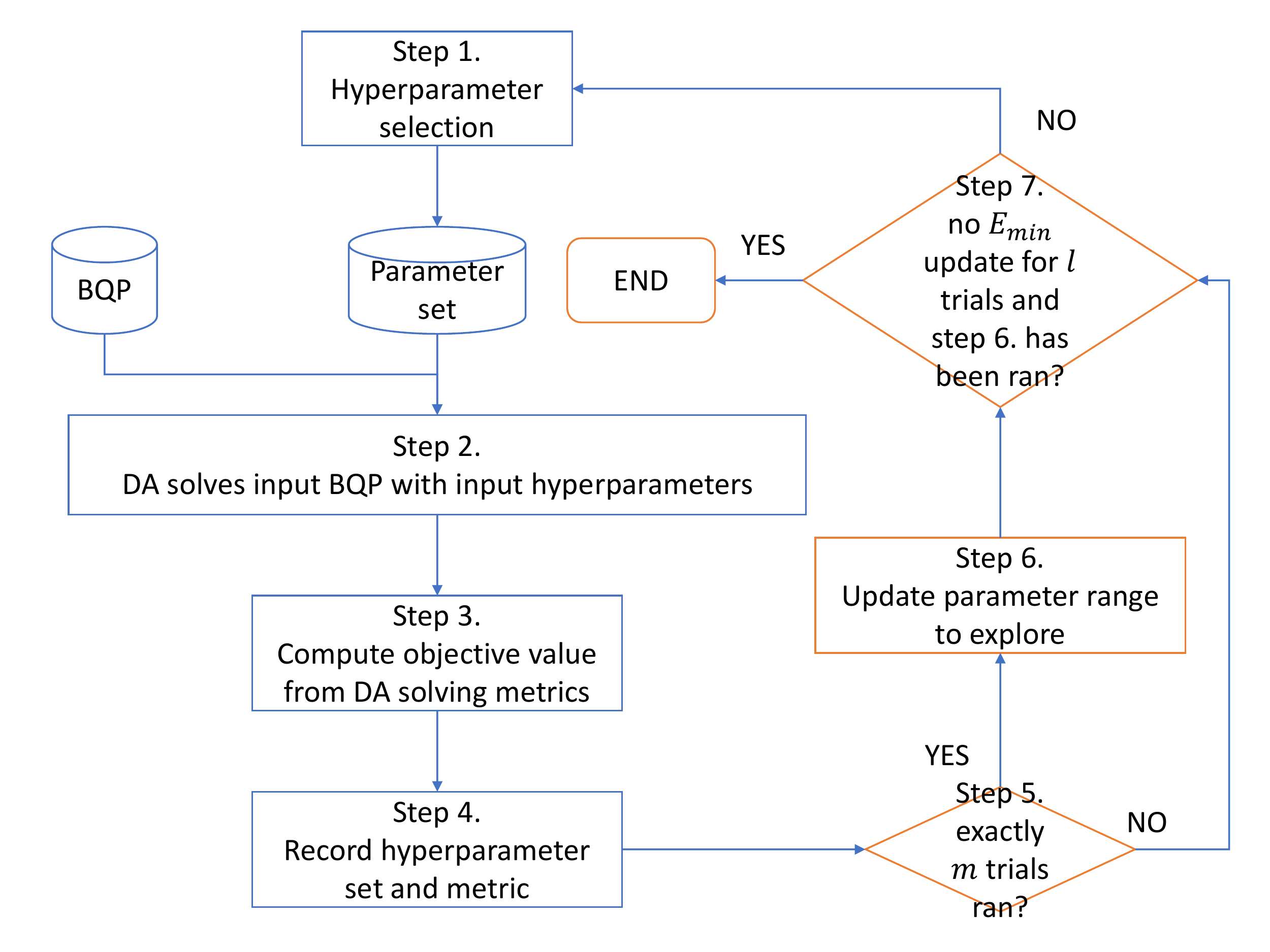}
\caption{DA fast-convergence tuning method}
\label{fig:fast_tuning}
\end{figure}

Our main contribution is a ``fast convergence" method to accelerate TPE when using TPE for optimizing hyperparameter values of Ising machines.
We describe the method in Fig. \ref{fig:fast_tuning}.

The differences with the method described in Fig. \ref{fig:tuning_flow} are twofold as we introduce:
\begin{itemize}
    \item \textbf{Range narrowing}: a tuning warm-up phase which lasts $m$ trials after which we update each hyperparameter range to explore (step 5 and 6 in Fig. \ref{fig:fast_tuning}).
    \item \textbf{Convergence judgment}: if $E_{min}$ was not updated for $l$ number of trials (step 7 in Fig. \ref{fig:fast_tuning}) after warm-up phase has been done, we terminate the tuning regardless of how many trials were left to be run in regard to set $n$.
\end{itemize}

\subsection{Range Narrowing}\label{sec:range_narrowing}
In step 6 in Fig. \ref{fig:fast_tuning}, for each parameter $x$ we update its search range $r$ by narrowing it and centering on current best parameter value. 
Thus $r_{new} = r_{old}/\gamma$ with $r_{new}$ the new range of a given hyperparameter, $r_{old}$ its previous range until $m$ trials were completed and $\gamma$ an introduced hyperparameter for our method. 
We note $m$ is also a newly introduced hyperparameter representing how long the warm-up phase will be. 
The new min and max of each parameter range is centered on their bast values after $m$ trials, they become $r_{{new}_{min}} = x_b- \frac{r_{new}}{2}$ and $r_{{new}_{max}} = x_b + \frac{r_{new}}{2}$ where $x_b$ is the parameter value leading to $E_{min}$ during the warm-up phase.
If $r_{{new}_{min}}$ would become lower than the lowest value allowed for that parameter, it is clipped to to this lowest allowed value.
If $r_{{new}_{max}}$ would become higher than the highest value allowed for that parameter, it is clipped to to this highest allowed value.

The intuition behind our concept of range narrowing is that hyperparameter range allowed by an Ising machine such as DA as its input can be significantly wide and the actual ``effective" range can be much narrower. 
Thus, even when using intelligent method such as TPE, it can be challenging and our proposed range narrowing method can be effective although we introduce two new hyperparameters in our method: $m$ and $\gamma$.

\subsection{Convergence Judgment}\label{sec:conv_judg}
Step 7 in Fig. \ref{fig:fast_tuning} role is to avoid doing hyperparameter trials which have a low probability of finding better hyperparameter values and thus lowering $E_{min}$ when a significant number of trials has already been completed and $E_{min}$ has not been updated for a set number of trials $l$.
To implement this part, after warm-up phase is over, we simply count how many trials have been done without updating $E_{min}$. 
This number of trials without $E_{min}$ update is represented by $l$ in step 7 in Fig. \ref{fig:fast_tuning}.
If more than $l$ trials have been completed without updating $E_{min}$, the tuning is terminated.

\subsection{Newly Introduced Hyperparameters}\label{sec:new_hyper_param}
As we we have described in Section \ref{sec:range_narrowing} and \ref{sec:conv_judg}, we have introduced three new hyperparameters: $l$, $m$ and $\gamma$.
In this sub-section we discuss the trade-off of introducing new hyperparameters, with their related complexity, against expected benefit of finding better parameters, faster.

First, we make the claim that $l$ virtually replaces $n$ and is easier to tune.
Our reason is, now instead of setting a fixed number of trials to run $n$, hoping it will be large enough to reach low enough $E_{min}$ but not so large that a significant amount will be ran without any improvement, and thus wasting Ising machine time, we have $l$ where the user consciously decides after how many trials without $E_{min}$ updates he judges the hyperparameter tuning should end.
The user should set $l$ proportionally to the BQP difficulty he wants to solve, with a larger $l$ value for more difficult problems. 
Thus, we think our proposed convergence judgment should not just benefit TPE but any black-box tuning method where a certain number of trials to execute has to be input.

For $m$, the length of our warm-up phase for hyperparameter range narrowing described in Section \ref{sec:range_narrowing}, like for $l$, we suggest to set it at a value proportional to BQP difficulty to solve. We also note it has to be inferior to $n$.
We suggest to have $m=l$, as we used in Section \ref{sec:experiments}.

For $\gamma$, we suggest to use a value of 4, as reducing the $r$ to a fourth of its width worked well in our experiments in Seciton \ref{sec:experiments}.
A sidenote on BQP difficulty, BQP size can be used as a proxy as well as the number of linear inequalities, but other more sophisticated approaches can also be considered.

\section{Experimental Results}\label{sec:experiments}
As we stated in Section \ref{sec:hyper_param_imp}, we focus on BQP difficult enough so that when we try to solve them using DA within run time $T$, we will never reach its optimal solution, no matter what parameter was used.
We chose two well-known permutation problems Travelling Salesman Problem (TSP) and Quadratic Assignment Problem (QAP).
Within their publicly available benchmark, respectively ``tsplib"\cite{tsplib} and ``qaplib"\cite{qaplib}, we chose 2 instances from each.
First, we will describe TSP and QAP formulations as BQP.

\subsection{Travel Salesman Problem as a Binary Quadratic Program}
A well known formulation of Travel Salesman Problem as QUBO is described in \cite{lucas_karp}, the cost of the QUBO is:
\begin{alignat}{2}
\label{eq:tsp_qubo_cost}
    E(x) = \sum_{i=1}^{N}\sum_{j=1}^{N} D_{i,j}x_{i,t}x_{j,t+1}
\end{alignat}
\noindent where $D$ is the TSP distance matrix, $N$ is the number of nodes of the TSP, $x_{i,t}$ is the $i$ visited node at time slot $t$, $x_{j,t+1}$ is the $j$ visited node at next time slot $t+1$.
We thus have $N$ binary variables per time slot which represent for each time slot, which node was visited, for a total of $N^2$ binary variables. 
The goal is to find the order in which to travel each node so that \eqref{eq:tsp_qubo_cost} is minimized.
The coefficients between each binary variable constitute $Q_{obj}$, the objective input of the DA.
The constraints of the TSP are that each node should be visited once and only once and that at each time slot, only one node should be visited. 
This can be expressed as:
\begin{alignat}{2}
\label{eq:tsp_qubo_const}
     H_{pen}(x) = \sum_{t=1}^{N}\Bigg(1-\sum_{i=1}^{N}x_{i,t}\Bigg)^2 + \sum_{i=1}^{N}\Bigg(1-\sum_{t=1}^{N}x_{i,t}\Bigg)^2
\end{alignat}
\noindent The coefficients between each binary variable in $H_{pen}(x)$ constitute $Q_{pen}$, the constraint QUBO input of the DA. If both constraints are respected, $H_{pen}(x)$ value will be 0.

\subsection{Quadratic Assignment Problem as Binary Quadratic Program}
A well-known formulation of QAP as a QUBO has been established in \cite{glover_tutorial_qubo_qap}. 
From it, we derive its formulation as BQP:
\begin{alignat}{2}
\label{eq:qap_qubo_cost}
    E(x) = \sum_{i=1}^{N}\sum_{j=1}^{N}\sum_{k=1}^{N}\sum_{l=1}^{N} F_{i,j}D_{k,l}x_{i,k}x_{j,l}
\end{alignat}
\noindent where $x_{i,k}$ is a binary variable which represents factory $i$ assigned at location $k$, $F$ is known as the flow matrix, representing the amount of exchange between each factory, and $D$ the distance matrix between each location.
We thus have $N$ binary variables per location which represent for each location, which factory should be assigned, for a total of $N^2$ binary variables.
The goal is to assign each factories at locations such that \eqref{eq:qap_qubo_cost} is minimized.
The coefficients between each binary variable constitute $Q_{obj}$, the objective input of the DA.

The constraints are similar to TSP, each factory should be assigned once and only once, and at each location only one factory should be assigned:
\begin{alignat}{2}
\label{eq:qap_qubo_const}
     H_{pen}(x) = \sum_{i=1}^{N}\Bigg(1-\sum_{j=1}^{N}x_{i,j}\Bigg)^2 + \sum_{j=1}^{N}\Bigg(1-\sum_{i=1}^{N}x_{i,j}\Bigg)^2
\end{alignat}
\noindent likewise the coefficients between each binary variable in $H_{pen}(x)$ constitute $Q_{pen}$, the constraint QUBO input of the DA. If both constraints are respected, $H_{pen}(x)$ value will be 0.

\subsection{Experimental Settings}\label{sec:exp_settings}
We chose two difficult instances from tsplib and qaplib: respectively kroA100, gr120 for TSP and tai80a, tai100a for QAP.
We run one experiment per problem instance. We describe the common settings between all experiment (and thus all problem instances BQP) below.

We used Optuna which allows to choose between both random and TPE sampler. 
Among the hyperparameters available for the DA, we chose to tune ``gs\_level", ``gs\_cutoff", ``num\_run" and ``num\_group" as they are the only parameters related to the search engine performance. Their description available in \cite{gs_level_setting} is:
\begin{description}
\item[num\_run:] The number of parallel attempts of each groups (int64 type).
num\_run x num\_group specifies the number of parallel attempts.

\item[num\_group:] The number of groups of parallel attempts (int64 type).
num\_run x num\_group specifies the number of parallel attempts.

\item[gs\_level:] Level of the global search (int64 type).
In the global search, the search starting point with local solution group escape is determined, and the constrained search combining various search methods is repeatedly executed as a processing unit. The higher the value, the longer the constraint exploitation search.
Specifies the level of the global search. Lower level is weak on Global Search.

\item[gs\_cutoff:] Global search cutoff level (int64 type).
Specifies the convergence judgment level for global search constraint usage search. The higher the value, the longer the period during which the constraint-based search energy on which convergence is based is not updated. Convergence assessment is turned off at 0.
\end{description}

We tuned the above parameters using their full parameter range, ``gs\_level": $[0, 100]$, ``gs\_cutoff": $[0, 10^6]$, ``num\_run": $[1, 16]$, ``num\_group": $[1, 16]$. 

Those parameters are integers, thus the total search space is approximately $25.9\times 10^9$ combinations of hyperparameter values. 
For each trial, we use a DA run time $T$ of 30 seconds, exploring the full space would require approximately twenty-five thousand years.
We run $n=1000$ trials. We use $\gamma =4$ to reduce the range of parameters to a fourth of their original range after the warm-up phase is completed, as described in Section \ref{sec:range_narrowing}.
We compare, after each trial, each baseline method and the proposed fast-convergence method average cost value of their best-found feasible solution, for their best-found hyperparameter set since the start of the experiment over 5 different Optuna random seeds. 
We plot them respectively on x and y axis in Fig. \ref{fig:kroA100}, \ref{fig:gr120}, \ref{fig:tai80a} and \ref{fig:tai100a}. 
In addition, we also plot as a horizontal line the cost obtained after $T=30s$ using default hyperparameter values.

\subsection{Results}\label{sec:results}
\begin{figure}[tb] 
\centering
\includegraphics[width=0.6\columnwidth]{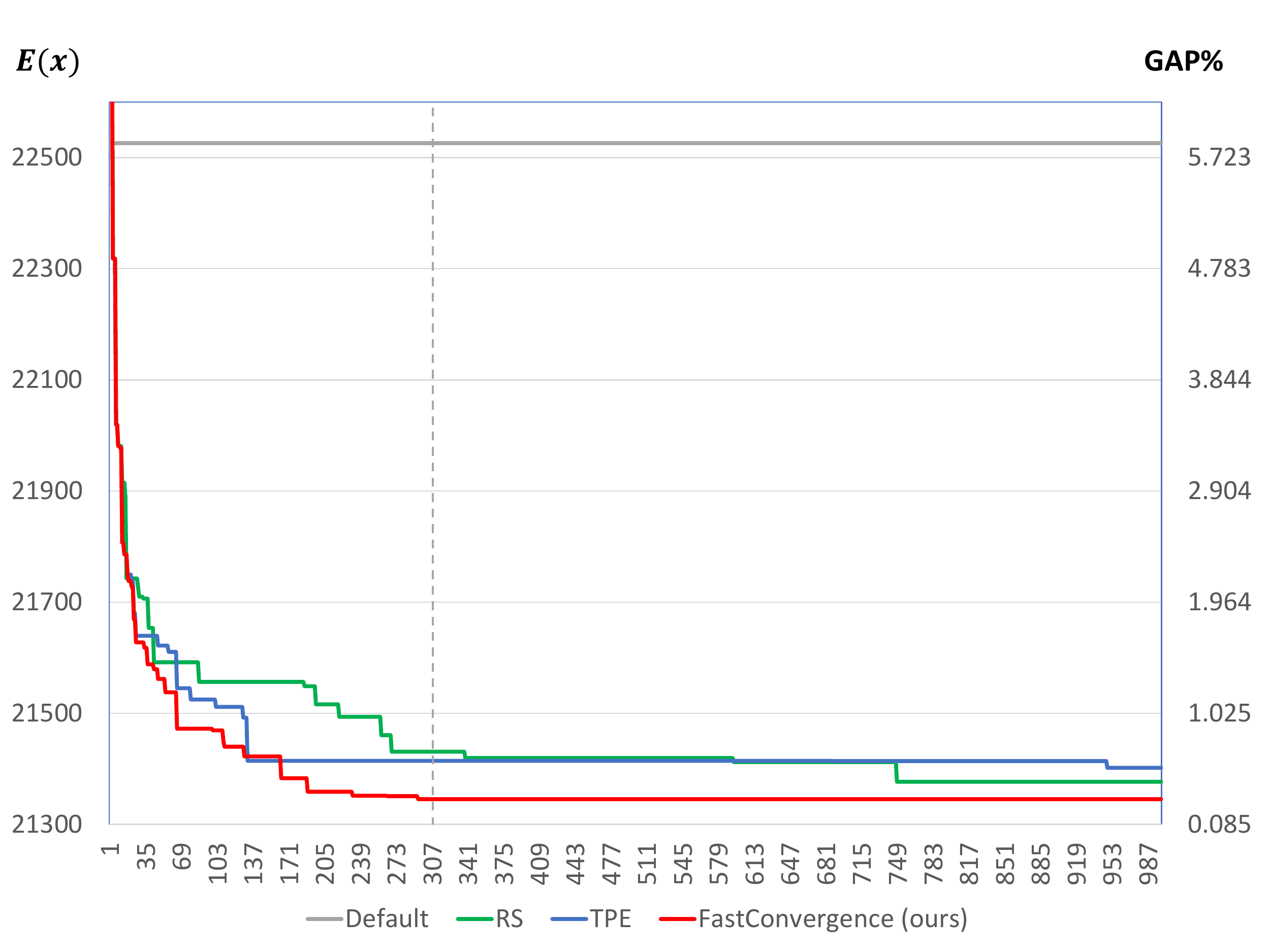}
\caption{TSP kroA100 Results}
\label{fig:kroA100}
\end{figure}

For KroA100 instance, which is composed of 100 nodes, we use $l=m=150$ as settings for step 5 and 7 of our FastConvergence method in Fig. \ref{fig:fast_tuning}.
We then observe the results in Fig. \ref{fig:kroA100}.
Gray line represents the cost value when using default hyperparameters values for $T$, green line Random Sampling, blue line TPE and red line the proposed FastConvergence method.


We observe FastConvergence performs better than random sampling and TPE.
It converges in average after 304 trials, referred to as "END" on Fig. \ref{fig:fast_tuning} and is symbolized by the grey dashed line in Fig .\ref{fig:kroA100}. 
Whereas for other methods, $n=1000$ trials are ran.
At the 304 trials point, there is a GAP of 0.3\% in favor of FastConvergence against TPE, where GAP means the relative difference in $E(x)$ between FastConvergence and TPE.
This instance illustrates how if the total number of trials  $n$ to complete tuning was smaller than 150, performance would have been significantly worse and how going beyond 150 trials yields only marginal improvements.
Thus, FastConvergence yields top performing parameters for approximately a third of the tuning time of standalone TPE and random sampling.


\begin{figure}[tb] 
\centering
\includegraphics[width=0.6\columnwidth]{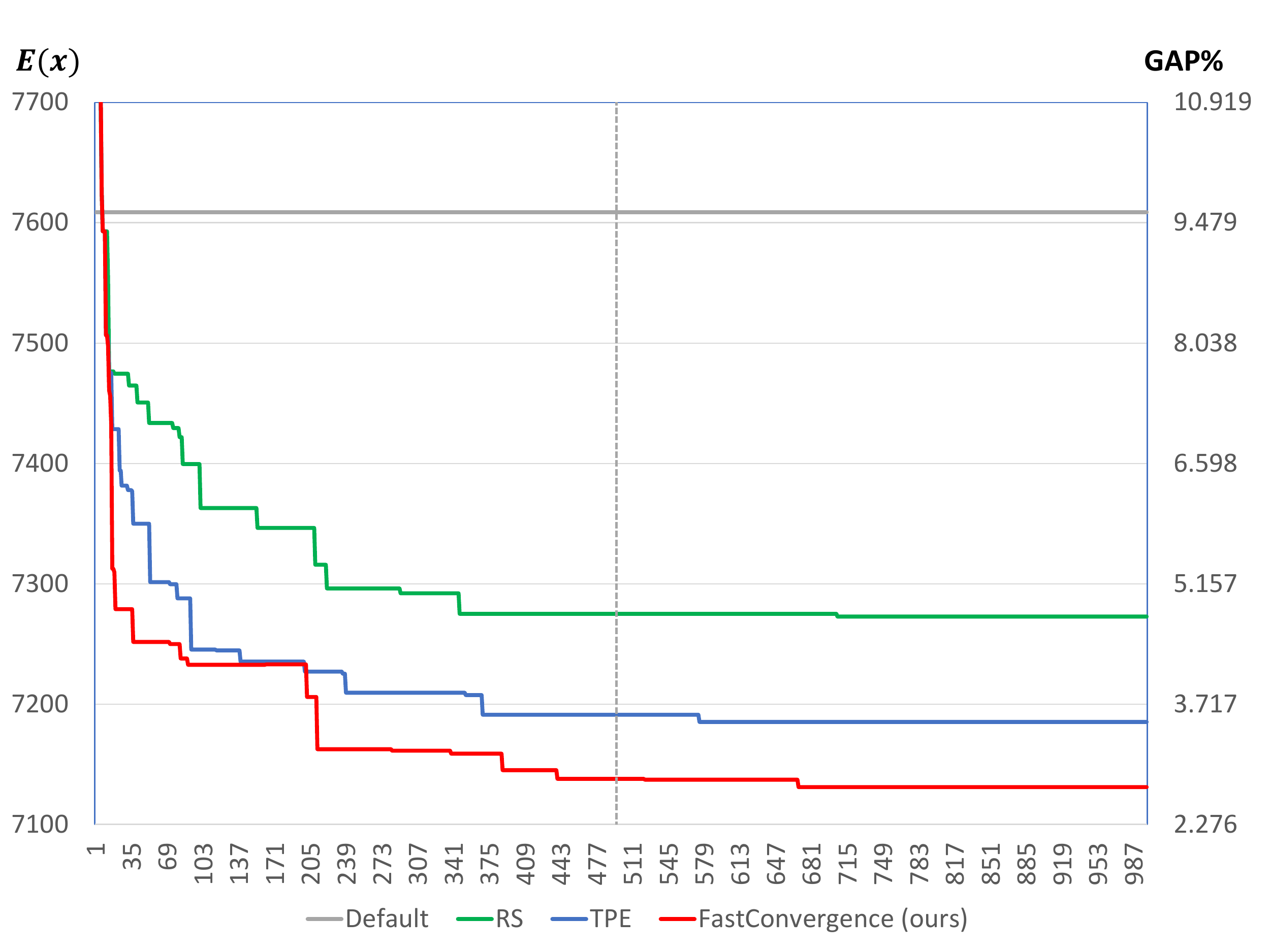}
\caption{TSP gr120 Results}
\label{fig:gr120}
\end{figure}

For TSP instance gr120, which is composed of 120 nodes, we use FastConvergence settings $m=l=200$ and observe the results in Fig. \ref{fig:gr120}.
We setup a higher value for $l$ and $m$ since the number of TSP nodes is larger.
In this instance, compared to kroA100, random sampling perform significantly worse than the other two methods.
We note FastConvergence converges after 496 trials in average, at which point the GAP with TPE is of 0.7\% in favor  FastConvergence which is larger than for kroA100.

\begin{figure}[tb] 
\centering
\includegraphics[width=0.6\columnwidth]{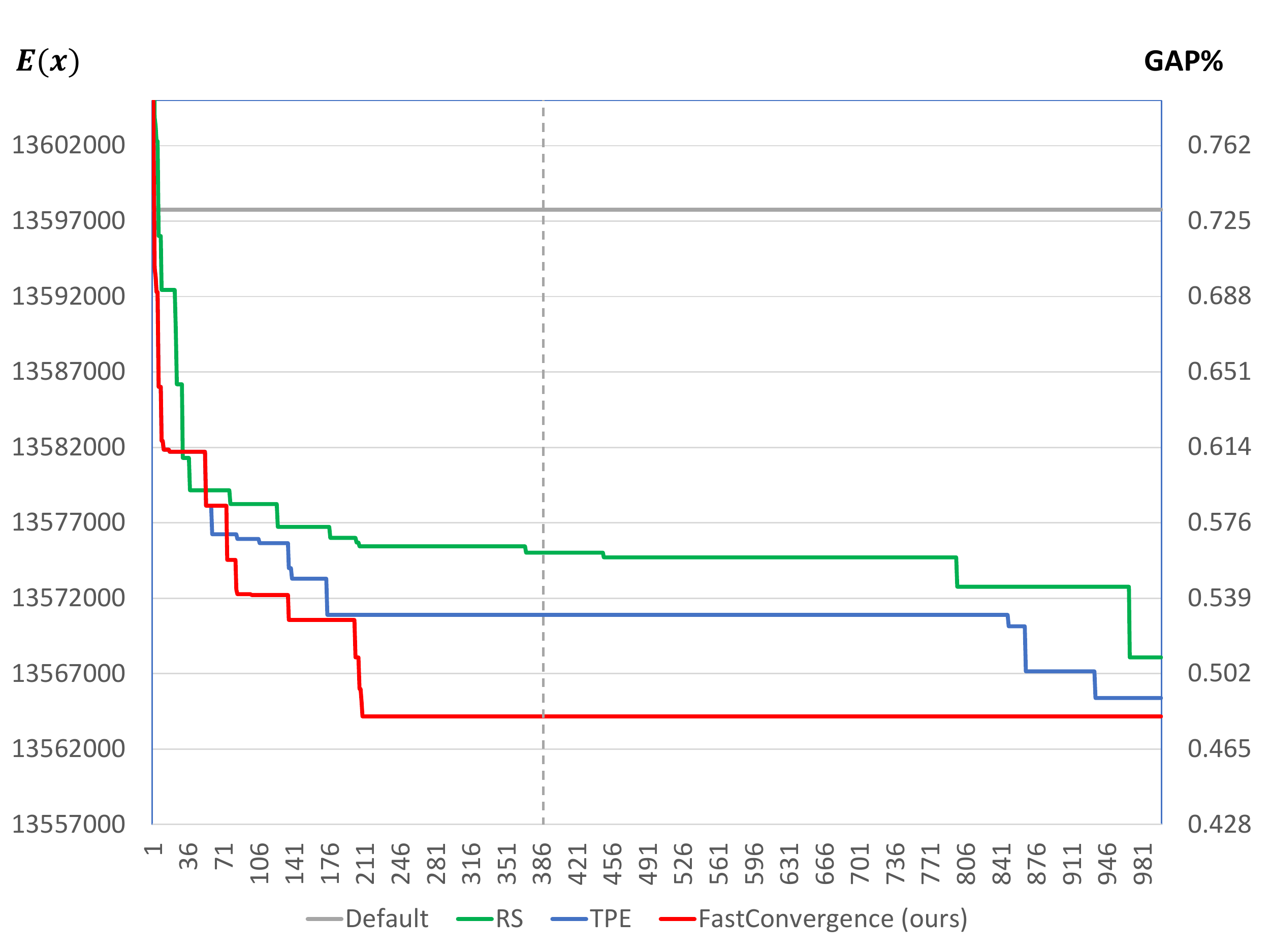}
\caption{QAP tai80a Results}
\label{fig:tai80a}
\end{figure}

For QAP instance tai80a, which is composed of 80 factories and locations, we use FastConvergence settings $m=l=200$ and observe the results in Fig. \ref{fig:tai80a}.
Although the number of variables is smaller than for gr120, we setup a similar value to gr120 for $l$ and $m$ since QAP is a more complex problem than TSP.
We still have FastConvergence, TPE, random sampling in order of most well performing tuning technique as for TSP.
FastConvergence converges in average after 388 trials at which point the GAP with TPE is 0.05\%.

\begin{figure}[tb] 
\centering
\includegraphics[width=0.6\columnwidth]{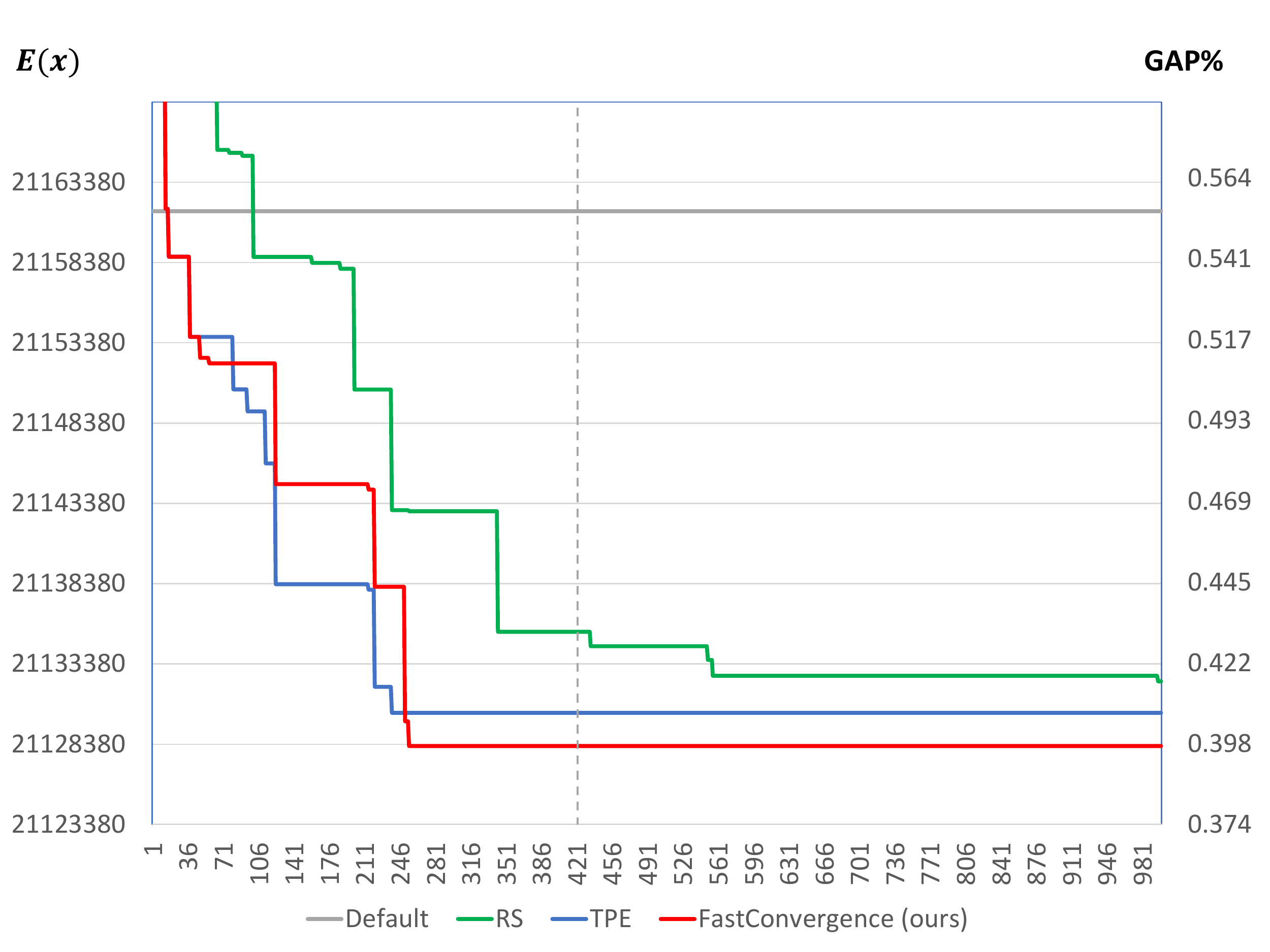}
\caption{QAP tai100a Results}
\label{fig:tai100a}
\end{figure}

For QAP instance tai100a, which is composed of 100 factories and locations, we use FastConvergence settings $m=l=250$ and observe the results in Fig. \ref{fig:tai100a}.
FastConvergence converges after 422 trials in average at which point the GAP with TPE is 0.01\% and is the best performing method overall.

Across all tested instances, FastConvergence was able to find better parameters than TPE alone, within between 304 and 496 trials. Compared to the $n=1000$ trials, it represents approximately between a third and a half of the learning time to obtain similar performing  hyperparameter values.

\section{Conclusion}\label{sec:conclusion}
In this paper, we showed popular black-box tuning techniques such as random sampling or tree-structured Parzen estimator can be used  effectively to tune Ising machines. 
We proposed to improve on the state of the art, TPE, to allow it to converge faster while still being able to obtain similar parameter quality.
The proposed FastConvergence method as well as the objective we defined for TPE are the first steppingstones in our hyperparameter tuning framework for Ising machines.

In our futures works, we would like to try FastConvergence on other problem categories and Ising machines to demonstrate how general it is. 
We also would like to study how hyperparameter values in general can be re-used across different problem instance and how we could do that efficiently to accelerate tuning time.

\bibliographystyle{IEEEtran}
\bibliography{bibliography}

\end{document}